# Task Allocation and Coordinated Motion Planning for Autonomous Multi-Robot Optical Inspection Systems

Yinhua Liu[1], Wenzheng Zhao[1], Tim Lutz[2], Xiaowei Yue[2,*]

**Abstract:** Autonomous multi-robot optical inspection systems are increasingly applied for obtaining inline measurements in process monitoring and quality control. Numerous methods for path planning and robotic coordination have been developed for static and dynamic environments and applied to different fields. However, these approaches may not work for the autonomous multi-robot optical inspection system due to fast computation requirements of inline optimization, unique characteristics on robotic end-effector orientations, and complex large-scale free-form product surfaces. This paper proposes a novel task allocation methodology for coordinated motion planning of multi-robot inspection. Specifically, (1) a local robust inspection task allocation is proposed to achieve efficient and well-balanced measurement assignment among robots; (2) collision-free path planning and coordinated motion planning are developed via dynamic searching in robotic coordinate space and perturbation of probe poses or local paths in the conflicting robots. A case study shows that the proposed approach can mitigate the risk of collisions between robots and environments, resolve conflicts among robots, and reduce the inspection cycle time significantly and consistently.

## 1. Introduction

In recent years, autonomous robotics-guided optical measurement probes have been introduced to improve production inspection and inline quality control. Unlike traditional offline gauges such as coordinate measuring machines, non-contact robotics-guided optical probes have enabled inline measurement of highly complex surfaces in production. The obtained data can be used for real-time process monitoring, fault diagnosis, and in-process quality improvement. Due to the complexity of large-scale free-form surfaces in some products (e.g., auto bodies, aircraft), multiple industrial robots equipped with optical probes are usually distributed in the inspection station. This poses a multi-robot coordination problem. To increase the efficiency of quality inspection and mitigate conflict risks among multiple robots, we develop a rigorous task allocation and multi-robot motion planning methodology. The inline optical inspection system can accurately measure raw structures, sub-assembly parts, and final products in the flexible assembly line. Our methodology has the potential to advance the autonomy, safety, and efficiency of inline multi-robot coordination and real-time quality inspection.

We start the literature review with collision-free path planning for robots. Traditional single-robot collision-free motion planning methods include randomized kinodynamic planning (Kazemi et al., 2013), reinforcement learning (Hua et al., 2021), Neural Network (Pashkevich & Kazheunikau, 2005), and sampling-based planning method (Jaillet et al., 2010), etc. For a detailed review of path planning and optimization of mobile robots, we refer the reader to (Zafar & Mohanta, 2018). The aforementioned methods address path planning and optimization problems in static and dynamic environments and have been applied to different fields. However, these methods usually consider gently curved response surfaces.

With the advancement of product design and manufacturing technologies, large-scale and free-form surfaces have become more common, which raises new challenges for robotic path planning. For example, the latest auto body design may include various features such as holes, slots, surfaces with large curvature, edges, and threads. The existing methods mainly model the robot and its working environment, and then use heuristic algorithms such as the Artificial Potential Field method to search for a collision-free path of the robot. However, given the diverse features in large-scale free-form surfaces and a large number of joints in a typical industrial robot, it is very challenging for heuristic algorithms to find a feasible path without collision in real time. One common way to avoid a collision is to define intermediate transition points when trying to generate a collision-free path. Yet, these transition points are primarily derived from engineering experience, which is not captured by

* Xiaowei Yue
E-mail: xwy@vt.edu

1 School of Mechanical Engineering, University of Shanghai for Science and Technology, Shanghai 200093, China.
2 Virginia Polytechnic Institute and State University, BlacksburgVA 24060, USA.

Yinhua Liu[1], Wenzheng Zhao[1], Tim Lutz[2], Xiaowei Yue[2,*]

automatic programming and may not yield the shortest path. For 3D scanning inspection of free-form surfaces, Glorieux et al. (2020) proposed a target viewpoint sampling strategy to plan the coverage path to achieve minimum viewpoints and short trajectory of a single robot. This approach generated coverage paths with a shorter cycle time. Liu et al. (2020) developed an optimal path planning system for automated programming of measuring point inspection that incorporates probe rotations and effective collision detection. Although some efforts have been focused on this field, a complete task and motion planning strategy for multi-robot coordination and optimization still needs to be developed to improve the autonomous multi-robot optical inspection for complex free-form surfaces.

Approaches to multi-robot motion planning can be generally divided into two categories: centralized methods and decoupled methods (Clark, 2005). In the centralized methods, multiple robots working in a workspace are considered a single multi-body robot working in a composite, multi-degree-of-freedom configuration space. In the decoupling methods, a coordination phase is included during multi-robot motion planning. The decoupled methods have become more popular because of the smaller search space and higher calculation efficiency. To achieve the multi-robot coordinated motion, different strategies, including robot path changes, adjusting the robot speed, setting the robot priority, and delaying the startup (Chang et al., 1994; Bennewitz, et al., 2002; Zhong et al., 2014), were proposed. For the approaches above, a deadlock situation may happen because one robot (e.g., has a higher priority) blocks the path of another robot (e.g., has a lower priority). Some researchers proposed new methods to tackle this challenge. Nieto-Granda et al. (2018) proposed an adaptive informative sampling method for online coordination of multiple robots. Åblad et al. (2017) developed a surrogate model-based approach to partition the entire workspace and separate the workspace of each robot from the workspace of other robots to avoid conflicts. Dai et al. (2021) developed a framework to analyze the multi-robot coverage of large complex structures. Liu et al. (2014) proposed the dynamic priority-based path planning to address collision avoidance for the cooperation of multiple mobile robots.

With the integration of accessibility and workload balance among robots, task allocation also plays a critical role in multi-robot coordination. A good task allocation strategy can improve calculation efficiency and reduce the potential conflicts between robots, thereby reducing the cycle time in coordinated motion planning. In essence, the multi-robot task assignment problem is a Mixed Integer Linear Programming model (MILP), and it is an NP-hard problem. In practice, motion planning in an autonomous multi-robot optical inspection system demands an integrated solution for task assignment, robot path planning, and multi-robot coordination (Chakraborty et al., 2009; Spensieri et al., 2015). Some multi-robot motion planning approaches have specifically targeted applications in the manufacturing domain. For example, Wang et al. (2016) proposed a double global optimum genetic algorithm–particle swarm optimization to solve the welding robot path planning problem. Segeborn et al. (2014) developed a generalized simulation-based method for automatic robot line balancing to reduce the need for robot coordination by spatially separating the robot weld workloads. With a MILP model framework, the robotic welding manufacturing line was balanced considering the variation of robot parameters, distribution limits, motion time, and robot interference constraints (Lopes et al., 2017).

In summary, a task allocation strategy combined with a multi-robot coordination strategy that reduces conflicts is the key to ensure the high efficiency of a multi-robot inspection system. Although numerous studies have been conducted on the coordination among multiple robots, they cannot be applied to autonomous multi-robot inspection systems directly due to two aspects: (1) it is difficult to implement these approaches in complex large-scale free-form surfaces because the multi-robot optical inspection system has unique requirements on robotic end-effector orientations. (2) the optimization computation in existing approaches is time-consuming and cannot satisfy the inline inspection requirement. Furthermore, as the number of inspection tasks increases and the product geometry and industrial environment become more complex, the computation concerning reachability analysis, collision detection, and cycle time estimation will become even more complicated. This paper proposes a local robust inspection task allocation and a dynamic searching strategy for conflict avoidance to mitigate the risk of multi-robot conflicts and reduce the total cycle time of the inspection cell.

The remainder of this paper is organized as follows. Section 2 presents the physical modules and problem formulation of the autonomous multi-robot optical inspection system. In Section 3, a local robust task allocation method is proposed, which considers the distribution characteristics of measurement points and can realize efficient and well-balanced task allocation. In Section 4, a dynamic searching strategy is used to modify the joint configuration in the multi-robot coordinate space, which is represented as the position $(x, y, z)$ and orientation $\theta = (\alpha, \beta, \gamma)$ of a pose, of the conflicting robot to achieve the conflict-avoidance and coordinated motion planning. In Section 5, a case study of multi-robot optical inspection system on an auto body is presented to

Yinhua Liu[1], Wenzheng Zhao[1], Tim Lutz[2], Xiaowei Yue[2,*]

illustrate the effectiveness of the proposed strategy. The conclusions are summarized in Section 6.

## 2. Autonomous multi-robot optical inspection system

At present, auto-body inspection systems mainly employ three methods: offline gauges, three-coordinate measuring machines, and optical inspection systems. Among them, the throughput of offline gauges and three-coordinate measuring machines is low. Therefore, they are usually used for offline quality inspection rather than inline quality control in the fast-paced production line. Following recent advances, optical inspection equipment, such as a multi-robot optical inspection system, is becoming increasingly popular due to its high efficiency, inspection accuracy, and flexibility. This inspection method significantly reduces the flow time in multistage manufacturing and improves the performance of the quality inspection. Before developing our task allocation and coordinated motion planning methodologies for multiple robots, we firstly introduce the physical modules of this autonomous multi-robot optical inspection system and its characteristics. These characteristics will have a large impact on methodology development.

### 2.1. Optical Inspection Mechanism and Constraints

This section introduces how the optical inspection system works and what constraints it has. Fig. 1 shows one optical inspection system and the schematic diagram of its measurement. The non-contact probe uses an optical sensor based on the triangulation principle. The sensor projects a laser and a planar light simultaneously, which are used to obtain the accurate position of a measurement point (MP) in space according to the reflected light in the charge-coupled device camera imaging. Unlike 3D scanners that collect point cloud data offline, the new optical inspection gauge can measure the deviation of a single MP from each viewpoint of the probe in an inline setup. Usually, the autonomous multi-robot optical inspection system needs to obtain all the measurement points in a complex large-scale free-form surface in a short time. The MPs in this paper represent measuring characteristics located on free-form surfaces, such as holes, slots, surface points, and threads. Generally, the MPs are determined according to the assembly process design and quality requirements of customers, and the layout of the MPs is generated in the product design stage. The measurement data of the measured features in the sensor coordinate system can be calculated through the built-in feature fitting algorithm of the inspection gauges. The total cycle time of the auto body assembly line is 60 to 120 seconds. The optical inspection system should satisfy efficiency and accuracy requirements for inline quality inspection.

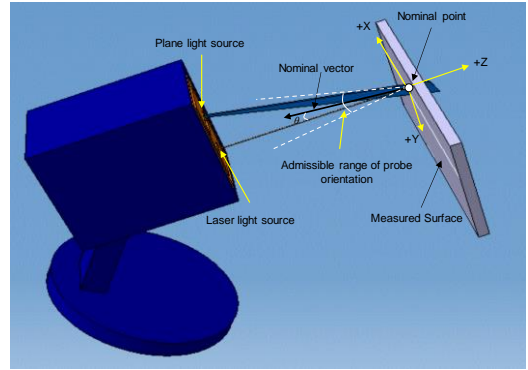

**Fig. 1.** One optical inspection system and the schematic diagram of its measurement

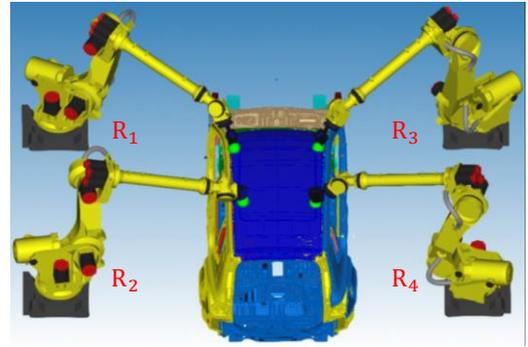

**Fig. 2.** An example of a multi-robot optical inspection station

In order to ensure inspection accuracy, several system constraints must be satisfied. Specifically, (1) the depth of field (DOF), which represents the depth from the position of the laser source to the MPs, must be within a predetermined range; (2) the MPs are required to be within the length of the laser stripe; (3) there is no interference between beams (e.g., incident beams and reflected beams) and the measured products and fixtures, etc.; (4) the view angle $\theta$, that represents the angle between the incident beam and the surface normal direction of the measured features, is supposed to be within a specified tolerance. For the best inspection precision, the view angle equals zero. In practice, the probe orientation is acceptable within $(-\theta_0, \theta_0)$, where $\theta_0$ is specified in the optical probe user manual.

### 2.2. Multi-Robot Optical Inspection System

We illustrate the multi-robot optical inspection process based on an automotive industry example. In an auto body assembly line, the multi-robot optical inspection system is used for the inline measurement of assembly deviations in critical MPs. Multiple MPs are located on the free-form surfaces of auto bodies. According to the previous section, the characteristics of optical gauges (e.g., the view angle) are required to be within a specified range of positions and orientations. The end-effectors, i.e., the optical probes, are mounted on the robots for dimensional inspection. The

Yinhua Liu[1], Wenzheng Zhao[1], Tim Lutz[2], Xiaowei Yue[2,*]

process of obtaining dimensional deviations at each MP can be regarded as one inspection task. In order to complete all the inspection tasks in the large-scale free-form surface efficiently, the inspection cell is generally composed of $m$ robots with optical probes mounted at the end of each robot. Fig. 2 shows a 4-robot inspection station that is commonly used in the auto body inspection process. In the limited cycle time of the assembly line, pre-specified critical MPs need to be measured by the multi-robot optical inspection system. For large-size and complex aircraft fuselage, the efficiency and quality precision requirements for optical inspection will also be high (Yue et al., 2018).

When we conduct multi-robot coordination and inspection optimization, MPs' position distribution and feature characteristics need to be considered. For example, an auto body product has large-scale and free-form surfaces. The MPs located on the auto body are numerous and various. These MPs have a variety of nominal directions. Many MPs, such as those located on the roof and middle area of the product, can be measured by more than one robot. Besides, considering the diverse nominal directions of MPs, poses of the end-effector and the robots' joints are required to change frequently during the inspection. We need to avoid the potential collisions between one robot and the product surfaces or fixtures, and meanwhile, we need to prevent conflicts among multiple robots. Therefore, an effective and efficient task allocation and multi-robot coordination strategy plays a vital role in collision avoidance and cycle time reduction in the inspection station.

### 2.3. Problem Formulation in Multi-Robot Optical Inspection System

This paper targets to reduce the cycle time by optimizing the task allocation, multi-robot coordination, and motion planning in the optical inspection system. To convey the problem formulation clearly, we summarize the assumptions in this study firstly.

The assumptions are pre-listed as follows.

1) When the optical probe reaches the position that meets the measurement specification of the optical inspection system, an MP is measured admissibly.

2) During the measurement process, each joint achieves its maximum velocity in a very short time.

3) The measuring time of each MP, excluding the time of the probe trajectory from one point to the other, does not change with the environment, i.e., the measuring time for each MP is consistent.

4) We do not consider the impact of failures of the robots during measurements.

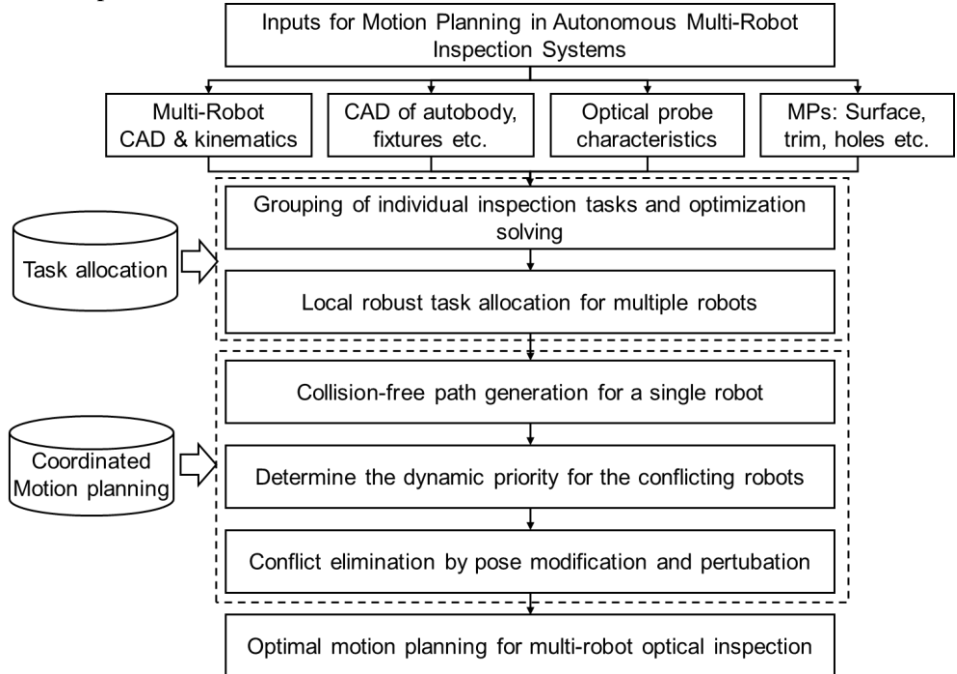

**Fig. 3.** Flow chart of the proposed method

The problem inputs include: 1) the multiple robot models and configurations, e.g., computer-aided design (CAD) geometries and their kinematic behaviors; 2) the CAD models of auto bodies and fixtures, and environment modeling; 3) the characteristics of the optical probe, DOF and the view angle range, etc.; 4) the set of inspection tasks located on the product surfaces, including spatial placements and surface orientations of the MPs. We assume MPs are given according to product design and quality requirements. The objective of the proposed research is to minimize the cycle time for accurate inspection as well as mitigate the risk of collisions in the

Yinhua Liu[1], Wenzheng Zhao[1], Tim Lutz[2], Xiaowei Yue[2,*]

inspection process. Based on the system configurations and problem statements, the following main procedures are performed: a) Considering the requirements of spatial position, normal direction of MPs and laser measurement characteristics, a local robust task allocation method based on a set cover optimization is proposed. The optimized task sets are assigned to each robot to reduce potential conflicts among multiple machines and improve the inspection efficiency; b) For a single robot with determined task allocation, the minimum inspection time is estimated based on the collision-free path planning algorithm; c) Without changing the initial sequence of motion planning, a coordination method by perturbing the orientation of the probe pose of the conflicting robot is proposed. Fig. 3 shows the flow chart of the method presented in this paper.

## 3. Local robust task allocation for multi-robot inspection optimization

### 3.1. Optimization for Multi-Robot Optical Inspection

The optical probes are mounted on the corresponding industrial manipulators/robots in the inspection cell. Multiple robots surround the product to be measured (e.g., auto body) and fixtures. Each robot is to start the inspection from its base position. After the robot completes all the assigned tasks, it will return to the original position and prepare for the following product inspection in the inline manufacturing line.

The purpose of the proposed method is to minimize the inspection cycle time for all MPs with different allocation schemes as well as to achieve the collision-free paths of multi-robot systems, given the constraint that the maximum time difference of multiple robots is less than $\gamma$. For all the allocation schemes, each MP should be measured only once. Assume that $\mathcal{R} = \{r_1, \ldots, r_m\}$ is the set of robots in the multi-robot inspection system; $\mathcal{H} = \{h_1, \ldots, h_n\}$ is the set of inspection tasks to be executed. The total number of robots is $m$ and the total number of tasks is $n$. An allocation $\mathcal{U} = \{U_1, \ldots, U_m\}$ is a task partitioning of the set $\mathcal{H}$ where $U_i$ is a sequence of tasks assigned to be executed by the robot $r_i$. The optimization problem for multi-robot optical inspection is represented as follows:

$$\min_{\mathcal{U}} \max_i T(r_i, U_i) \qquad (1)$$

$$\text{Subject to:} \max_{i,j} \left| T(r_i, U_i) - T(r_j, U_j) \right| \leq \gamma$$

$$\boldsymbol{G}(r_i) \cap \boldsymbol{V} = \emptyset$$
$$\boldsymbol{G}(r_i) \cap \boldsymbol{G}(r_j) = \emptyset$$
$$U_1 \cap U_2 \cap \cdots \cap U_m = \emptyset \qquad (2)$$
$$U_1 \cup U_2 \cup \cdots \cup U_m = \mathcal{H}$$
$$i, j = 1, 2, \ldots, m$$

where $\gamma$ represents the maximum acceptable difference of inspection time of different robots; The dynamic geometry space of the $i^{\text{th}}$ robot is denoted by $\boldsymbol{G}(r_i)$; The geometry space of the auto body and fixtures is denoted by $\boldsymbol{V}$; $T(r_i, U_i)$, or using $T_i$ for short, is the estimated motion time that the robot $r_i$ costs when executing the sequence of tasks $U_i$. For a defined order of the allocated tasks in $U_i$, the inspection time $T_i$ for the $i^{\text{th}}$ robot can be represented as

$$T_i = T_{n_i 0} + T_{01} + \sum_{j=1}^{n_i - 1} T_{j, j+1} \qquad (3)$$

where $T_{01}$ and $T_{n_i 0}$ represent the motion time of a robot from the initial position to the first MP and from the $n_i^{\text{th}}$ MP to the initial position, respectively; $T_{j, j+1}$ represents the time from the $j^{\text{th}}$ MP to the $(j+1)^{\text{th}}$ MP; $n_i$ is the number of MPs in $U_i$.

### 3.2. Local Robust Task Allocation for Multiple Robots

Given the large-scale free-form surface to be inspected, the MPs on the product, and configurations of multi-robot optical inspection system, we need to develop a task allocation for multiple robots. That is to determine the $U_i$ in Equation (1). Fig. 4. shows the layout of some MPs on an auto-body. Generally, there are multiple critical MPs located on the surfaces of an auto body. In Fig. 4, the red dots represent the nominal positions of the MPs, and the arrows indicate the surface orientations of the MPs.

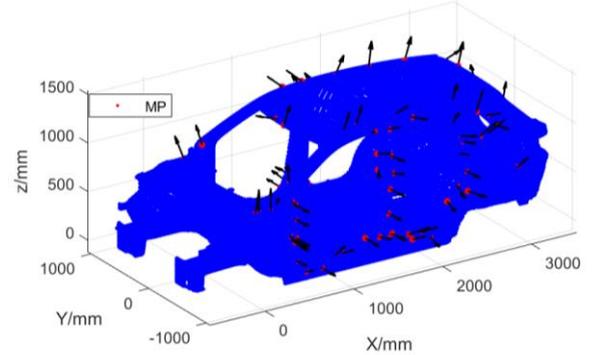

**Fig. 4.** Layout of MPs on an auto body surface

Due to limited reach, a single robot in the multi-robot optical inspection system cannot measure all the measuring points on the auto-body. Therefore, it is necessary to determine the working space of each robot. Generally, the robot's sampling of each joint angle can be realized by the Monte Carlo random sampling method. Then, the feasible area of the robot discretization can be determined by solving the robot kinematics problem.

Based on the working area of each robot, the proposed task allocation approach will distribute the MPs that two or more robots could measure in a manner that avoids collisions. The corresponding MPs that can only be measured by the $i^{\text{th}}$ robot is denoted by the set $\delta_i$, and the measuring points within this set that a single robot can measure are denoted as $q_k^i$. Some other MPs can be measured by multiple robots, and these MPs are denoted by

Yinhua Liu[1], Wenzheng Zhao[1], Tim Lutz[2], Xiaowei Yue[2,*]

$q'_j$. An effective task allocation method for $q'_j$ can not only reduce the motion time of each robot but also mitigate the risk of collisions among multiple robots.

To improve the inspection efficiency, the tasks $q'_j$ with short distance and similar probe poses will be allocated to the same robot. This idea is equivalent to introduce local robustness against small perturbations associated with locations and probe poses. This paper denotes the MPs that share a similar probe pose with a short distance as one MP set $\mathbf{S}_j$. When the same robot measures the features in this local robust MP set, it means a shorter trajectory of the robot and fewer probe poses change during the motion, which will reduce the probability of conflicts of multiple robots. We also represent the MPs that can be measured by more than one robot as a set $\mathbf{Q}_{il}$. Specifically, $\mathbf{Q}_{il}$ represents the set that can be measured by the $i^{\text{th}}$ robot and $l^{\text{th}}$ robot. $\mathbf{Q}_{il}$ may include multiple local robust MP sets. Fig. 5 shows an illustration of the MP sets.

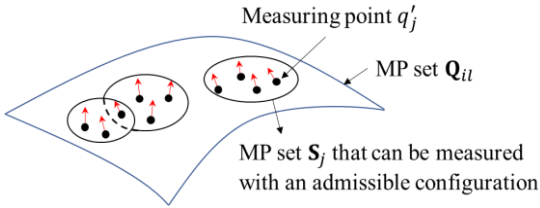

Measuring point $q'_j$

MP set $\mathbf{Q}_{il}$

MP set $\mathbf{S}_j$ that can be measured with an admissible configuration

**Fig. 5.** An illustration of the MP sets

The objective of task allocation is to distribute the MPs in $\mathbf{Q}_{il}$ to a specific robot so that total time for multi-robot inspection is minimized, and to do so in a computationally efficient manner. The procedure for the inspection task allocation is explained as follows.

- Step 1. Define the nominal vector of the probe and its specification for each $q'_j \in \mathbf{Q}_{il}$.

- Step 2. Determine multiple $q'_j$ for the local robust MP set $\mathbf{S}_j$ based on distance and probe orientation constraints.

- Step 3. Develop a set cover model for the local robust MP set selection.

- Step 4. Solve the set cover model to minimize the number of MP sets.

- Step 5. Assign the optimized MP sets to each robot.

- Step 6. Obtain task allocation results $U_i$ for the MPs.

Based on the allocated MPs $U_i$ for the $i^{\text{th}}$ robot, the procedure above can be repeated for different combinations of $\mathbf{Q}_{il}$ until all the MPs are allocated to a respective inspection robot.

We illustrate each step in detail. In Step 2, the MPs that meet the following conditions are added to the robust local sensor set $\mathbf{S}_j$.

$$\begin{cases} d(q_j, q_k) < \varepsilon \\ \theta_j \cap \theta_k \neq \emptyset \end{cases} \quad (4)$$

where $d$ is the distance between the nominal positions of different MPs, $\varepsilon$ is a threshold used for selecting adjacent MPs, and $\theta_j$ is the specification of probe orientation predetermined based on the optical inspection requirements. The introduction of local robust MP sets can deal with task allocation. A certain measure of robustness is sought against deterministic variability and uncertainty in the value of measuring locations and directions of optical probe poses.

In Step 3, given the MP sets defined based on (4), we can use the set cover to model the measuring task assignment for each $q'_j$. The minimum number of MP sets that cover all the inspection tasks in $\mathbf{Q}_{il}$ is needed. The set cover problem is a combinatorial optimization, and it is NP-hard. Usually, integer programming, the heuristic greedy algorithm, and the simulated annealing algorithm are used to solve the set cover problem. For a set cover with a large number of subsets, the computation based on integer programming is extremely expensive, and it is difficult for the heuristic optimization algorithms to obtain a stable globally optimal solution.

We can formulate the set cover problem into an unconstrained optimization problem through a penalty function. Firstly, the MP set $\mathbf{S}_j$ is transformed into an 0-1 matrix $a_{uj}$, and $a_{uj}$ indicates whether the cover set corresponding to the $j^{\text{th}}$ MP contains the $u^{\text{th}}$ MP or not. The objective function is to select the column from this matrix to minimize the cost of covering all rows, which can be expressed as:

$$min \sum_{j=1}^{n} c_j \beta_j$$

$$s.t. \begin{cases} \sum_{j=1}^{n} a_{uj} \beta_j = n, u = 1,2,\dots,n \\ \beta_j \in \{0,1\}, j = 1,2,\dots,n \end{cases} \quad (5)$$

where $c_j$ represents the shortest running time of the robot after that the $j^{\text{th}}$ column of measuring points is sorted; $\beta_j$ is a value 0 or 1 to represent whether the $j^{\text{th}}$ column is included in the solution. In order to meet the constraints, it is necessary to modify other elements in the row of element $a_{uj}$ to 0, that is, $a_{uk} = 0$ for $k = 1,2..n, k \neq j$.

Next, the problem is formulated into an unconstrained convex optimization problem by constructing a penalty function. The penalty function is $p(x) = n - \sum_{j=1}^{n} a_{uj}\beta_j$. Equation (5) is equivalent to Equation (6) below:

Yinhua Liu[1], Wenzheng Zhao[1], Tim Lutz[2], Xiaowei Yue[2,*]

$$\begin{cases} min\, f\,(x) = min\{\sum_{j=1}^{n} c_j\beta_j + \mu p(x)\} \\ \beta_j \in \{0,1\}, j = 1,2,\dots,n \end{cases} \quad (6)$$

where $\mu > 0$ is the penalty factor.

Then we use an auxiliary function that has the same discrete global minimum solution and solve a discrete dynamic convex optimization problem. The auxiliary function obtains the approximate discrete global minimum of the problem. For the situation above, the general search algorithm can obtain the discrete local minimizer solution $x_0$ and the current best discrete local minimizer $x_1^*$. To obtain the global minimum solution, the following auxiliary function is constructed.

$$T(x,k) = \begin{cases} f(x) + k||x - x_0||, if\, f(x) \ge f(x_1^*) \\ f(x), if\, f(x) < f(x_1^*) \end{cases} \quad (7)$$

where $k$ is a non-negative parameter and $|| \cdot ||$ indicates the penalty norm.

After the MP set extraction based on the set cover, the MP sets with the minimum set number will be assigned to each robot. The procedure for the MP sets allocation for multi-robot coordination is shown in Algorithm 1.

| Algorithm 1. Task allocation in multi-robot coordination |
|---|
| 1   **Input**: the *MP sets* $\mathbf{Q}_{il}$, the sets that can only be measured by the $i^{\text{th}}$ *or* $l^{\text{th}}$ robot, *i.e.* $\delta_i$ and $\delta_l$; $n_c$ is the number of subsets in $\mathbf{Q}_{il}$; $u_j$ represents the minimum subset of $\mathbf{Q}_{il}$ obtained based on convex optimization; $\varphi_i$ is the maximum MP number that can be allocated to the $i^{\text{th}}$ robot. |
| 2   **Output**: tasks assignment for the $i^{\text{th}}$ and $l^{\text{th}}$ robots, i.e. $U_i$ and $U_l$ |
| 3   **Initialization**: $U_i \leftarrow \delta_i$; $U_l \leftarrow \delta_l$ |
| 4   **For**   j=1:$n_c$ |
| 5     if    $n_i < \varphi_i$ |
| 6       Calculate the traveling distance $\text{Cost}_i$ established from the initial position $B_i$ and the set $u_j \cup U_i$, as shown in Equations (9) and (10). |
| 7     else |
| 8       $\text{Cost}_i = \infty$; |
| 9     end |
| 10    if    $n_l < \varphi_l$ |
| 11      Calculate the traveling distance $\text{Cost}_l$ established from the initial position $B_l$ and the set $u_j \cup U_l$, as shown in Equation (9) and (10) |
| 12    else |
| 13      $\text{Cost}_l = \infty$; |
| 14    end |
| 15    if    $\text{Cost}_i < \text{Cost}_l$ |
| 16      $U_i = u_j \cup U_i$ |
| 17    else |
| 18      $U_l = u_j \cup U_l$ |
| 19    end |
| 20   **end** |

As shown in Algorithm 1, the inputs include the MP sets $\mathbf{Q}_{il}$, the sets $\delta_i$ and $\delta_l$ that can only be measured by the $i^{\text{th}}$ or $l^{\text{th}}$ robot, the number of subsets in $\mathbf{Q}_{il}$, the maximum MP number $\varphi_i$ that can be allocated to the $i^{\text{th}}$ robot. The output is to complete tasks assignment for the $i^{\text{th}}$ and $l^{\text{th}}$ robots, i.e. $U_i$ and $U_l$. In each allocation, $n_i$ is the total number of MPs allocated to the $i^{\text{th}}$ robot. When the MPs $u_j$ is allocated to the $i^{\text{th}}$ robot, the cost is evaluated via the shortest distance among the MPs. The cost function is shown as follows.

$$\text{Cost}(U_i) = min\{d_0 + \sum_{k=1}^{n_i} \sum_{u=1}^{n_i} d_{ku}\alpha_{ku}\} \quad (8)$$

$$s.t. \begin{cases} \sum_{k=1}^{n_i} \alpha_{ku} = 1, u = 1,2,\dots,n_i \\ \sum_{u=1}^{n_i} \alpha_{ku} = 1, k = 1,2,\dots,n_i \\ \alpha_{ku} = \begin{cases} 1 \\ 0 \end{cases} \\ n_i \le \varphi_i \end{cases} \quad (9)$$

Yinhua Liu[1], Wenzheng Zhao[1], Tim Lutz[2], Xiaowei Yue[2,*]

where $d_0$ is the summary of the Euclidean distance from the base point to the first and last MP. $n_i$ is the number of MPs in the $i^{\text{th}}$ robot. $d_{ku}$ is the Euclidean distance from the $k^{\text{th}}$ MP to the $u^{\text{th}}$ MP.

## 4. Coordinated motion planning for multi-robot optical inspection

### 4.1. Collision-Free Path Planning for Individual Robot

After we complete the task assignment for each robot, the subsequent vital procedure is to plan a valid collision-free path for each robot. For this, methods such as grid-based method, genetic algorithm, reinforcement learning, heuristic optimization methods, have been developed (Kapanoglu et al., 2012; Zafar & Mohanta, 2018; Hua et al., 2021). The traditional methods usually determine a collision-free configuration space (C-Free space) of each robot, considering the robot geometry and the static environment. Then, a heuristic algorithm, such as the A* algorithm (Zafar & Mohanta, 2018; Zuo et al., 2015), is used to search the robot's collision-free configuration space and find an optimal path for each robot. However, in the presence of free-form surfaces and diverse surface orientations of a given product design, existing inspection approaches may not capture critical features in the curvature of product surfaces (Liu et al., 2020).

We have developed an optimal path planning approach for automated inspection programming via a single robot (Liu et al., 2020). This approach incorporated probe rotations and effective collision detection. Specifically, the structure of complex product surfaces can be discretized as a point cloud dataset, and a dynamic searching volume-based algorithm was proposed to detect potential collisions. A local path generation method was developed with the integration of the probe trajectory and probe rotation. By adding collision avoidance points properly, the collision-free inspection path between MPs could be determined. Then the motion time of each local path could be estimated. We developed an optimization approach to identify the global inspection path for all critical points and minimize the total inspection time. To realize better convergence and computation efficiency, the simulated annealing algorithm was used to solve the optimization problem (Aarts et al., 2003) .

### 4.2. Motion Planning for Autonomous Multi-Robot Coordination

The collision-free path planning of the multi-robot optical inspection pair mainly needs to meet the following criteria: *i)* when a single robot measures the assigned MPs within the specified time, there is no conflict between the robot and the auto-body, fixtures, etc. *ii)* there is no conflict between multiple robots, and meanwhile, the multi-robot operations should be as parallel as possible to improve the efficiency of the multi-robot optical inspection system. In the preceding section, we obtained the inspection task set $U_i$ of each robot and the collision-free path of a single robot. In Section 3, we realized the allocation of robot measurement tasks based on sensor measurement requirements and robot reachability constraints to ensure that the robot can measure the MPs with fewer pose changes and further reduce the probability of conflicts between multiple robots. At the same time, in Section 4.1, we obtained the collision-free path of a single robot.

Based on the procedure above, one robot may realize very efficient and accurate inspection. However, in multi-robot optical inspection, there is usually a shared space that multiple robots can reach. When more than one robot reaches this space at the same time, collisions among robots may occur. Provided with the outcome of each robot's task allocation and collision-free path planning, we will develop a coordinated motion planning approach to eliminate the potential conflicts. Firstly, the priority of the conflicted robots is determined based on the inspection time of the path planned for the task set $U_i$. The shorter the motion time of one robot is, the higher a priority we assign to this robot. Then, for the conflicting local path between the MPs to be serially inspected, the joint configuration for the robot with higher priority will be modified. In this way, the inspection efficiency can be improved, and the collision risk can be mitigated.

For a fixed position of the robotic end-effector, there will be numerous poses for the inspection robot to reach a target position. While, even the position of the end-effector remains, the trajectory of the local path between different measurement points may be changed by adjusting the postures of the robots during the measurement. The collision-free state space considering the dynamic robots and the static environment is denoted by $\Gamma = (x, y, z, roll, pitch, yaw)$. With a fixed placement of the robotic end-effector, a new strategy by modifying $(roll, pitch, yaw)$ in $\Gamma$ is proposed to eliminate the robots' conflicts within the given specifications of optical probe orientations.

Based on the position where one conflict may occur, the collision between robots can be categorized into two scenarios. Scenario (1): two robots collide when their optical probes are measuring the point $p_j$. Scenario (2): the robots collide when their optical probes are moving from the previous MP to the next one. i.e. in the middle of the path. Scenario (1) may occur with very low probability. In this scenario, the measuring pose of the robot with higher priority will be modified automatically within the probe pose specifications. For Scenario (2), the goal for conflict elimination can be achieved by discretizing the acceptable space of the probe based on the optical probe characteristics. Then, under the condition of no collision,

Yinhua Liu[1], Wenzheng Zhao[1], Tim Lutz[2], Xiaowei Yue[2,*]

the discrete T-space, which indicates the end-point coordinates $(x, y, z)$ of one robot, will be searched. Next, the probe poses with the shortest motion time from $p_j$ to the next MP will be selected to eliminate conflicts.

- Step 1. For the $i^{th}$ and $l^{th}$ robot that may have conflicts, determine the priority of robots. (We assume $T_i < T_l$ when we illustrate Step 2-4).
- Step 2: Find the measurement points before and after a potential collision in the $i^{th}$ robot, and denote them as $p_a$ and $p_b$. The feasible inspection poses are determined by perturbing the orientation of the probe: $(w, p, r) = (w_0, p_0, r_0) + (\Delta w, \Delta p, \Delta r)$; $(w_0, p_0, r_0)$ is the orientation of the robot when measuring $p_a$ or $p_b$; $(\Delta w, \Delta p, \Delta r)$ indicates an increment.
- Step 3: The validity of $(w, p, r)$ is determined through the reachability and static collision detection between the robot and product surface. If no collision occurs, the orientation is considered valid; otherwise, invalid.
- Step 4: For the selected effective orientation, conflict detection is conducted on the path between the $i^{th}$ robot and the $l^{th}$ robot. Meanwhile, estimate the running time of the collision-free path of two robots, and select the orientation $(w, p, r)$ with the shortest running time.
- Step 5: Repeat Steps 1-4 for the next potential conflict.

## 5. Case study

To evaluate the effectiveness of the proposed method, we conducted a case study in an inline multi-robot inspection station for auto bodies. In the inspection cell, FANUC 2000iB-125L industrial robots are mounted with optical probes for dimensional deviation inspection. Four robots are denoted by $R_1$, $R_2$, $R_3$ and $R_4$, respectively. The locations of these sensing robots for the multi-robot optical inspection station are shown in Fig. 2. The layout of the optical sensing station for the left-side auto body, and the locations and surface normal directions of MPs are shown in Fig. 6. In Fig. 6(b), 45 MPs with different surface orientations, numbered from $M_1$ to $M_{45}$ are shown. The proposed method has been programmed to allocate the MPs, optimize the independent path of each robot, and finally modify the path to get an optimal collision-free coordination path. The parameters used in the case study are listed in Table 1.

The fixed priority method and the delay startup method were introduced for comparison analysis in multi-robot coordination (Shin & Zheng, 1992; Qadi & Goddard, 2005; Chen & Li, 2017). For the task allocation procedure, the principle of proximity (Wang et al., 2016; Eidenbenz & Locher, 2016) was used as a benchmark method. Firstly, we illustrate the proposed local robust task allocation method via the example of $R_1$ and $R_2$ robots. Assume the MP set $\{M_1, M_2, ..., M_{12}\}$ will be measured by $R_1$ and $R_2$. A local robust subset is determined based on the probe orientations and measuring distance for each MP in this set. Then, a set cover problem is solved to obtain the minimum number of MP sets. The minimum covering sets are listed as $\boldsymbol{S_j^*} = \{\{M_1, M_2, M_3\}, \{M_4, M_5, M_6\},$
$\{M_8, M_9, M_{12}\}, \{M_{10}, M_{11}\}, \{M_7\}\}$. Furthermore, according to MP allocation algorithm 1, the MPs assigned to $R_1$ and $R_2$ are $\boldsymbol{S_{R1}} = \{M_1, M_2, M_3, M_7, M_{10}, M_{11}\}$ and $\boldsymbol{S_{R2}} = \{M_4, M_5, M_6, M_8, M_9, M_{12}\}$, respectively. With similar procedures, the complete task allocation results based on the proposed method and the principle of proximity are shown in Table 2. The difference between the allocation results is highlighted in grey.

**Table 1** The parameters used in the case study

| Parameter | DOF | $\theta_0$ | $m$ | $\varphi_i$ | Inspection time per MP |
|-----------|-----|-----------|-----|------------|------------------------|
| value | 100mm | 15° | 4 | 14 | 1.5 |

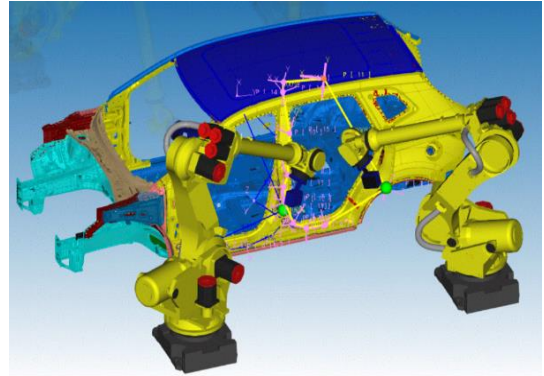

(a)

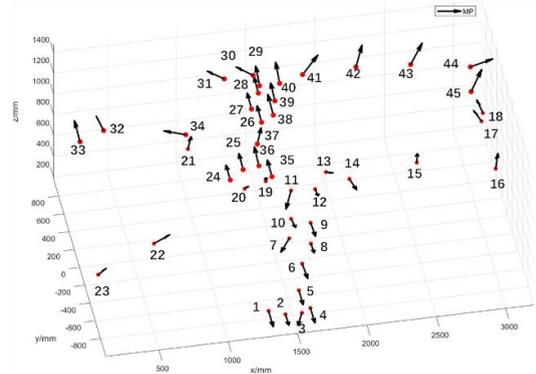

(b)

**Fig. 6.** (a) Layout of the multi-robot optical inspection system for left-side auto body; (b) Layout of MP locations on the auto body.

In the path planning for each robot, the collision detection algorithm was used to detect the collision between the robot and product/fixture surfaces. The automated inspection programming was used to determine

Yinhua Liu[1], Wenzheng Zhao[1], Tim Lutz[2], Xiaowei Yue[2,*]

the collision-free path for each robot. Next, the conflicts between robots were calculated based on the predetermined robot paths; the non-conflict paths were generated using the proposed multi-robot coordination strategy. Before the conflict elimination phase among robots, the robot time-position charts based on the principle of proximity and the proposed task allocation method are shown in Fig. 7 and Fig. 8, respectively. In the figures, $t$ represents the robot running time from the original position of each robot, and $p$ is the spatial coordinate of the probe central position when the robots collide. In Fig. 7, $R_1$ and $R_2$ collided when the time was at 3.9s and 5.9s, and $R_3$ and $R_4$ collided when the time was at 6.6s. In Fig. 8, $R_1$ and $R_2$ have a conflict at the time was at $t_2 = 7.9$s, $R_3$ and $R_4$ have a conflict at 23.7s. The red dash lines in the figures mean the conflicts of the robots. Different colors of the rectangles are used to distinguish the states before and after the conflicts of different robots.

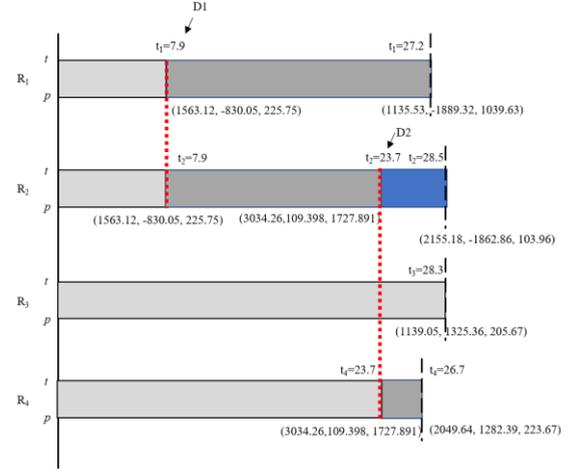

**Fig. 8.** Before multi-robot coordination, Time-position chart based on the proposed task allocation method (Red dash line represents the moment of conflicts between robots）

For performance comparison with other coordinated strategies for conflict elimination among robots, we used the cycle time that completes the measurement tasks without conflicts as the index. The delayed startup method determines the delayed time of the robot according to the conflicting time and locations. After the conflict robot passes through the collision position, the delayed robot starts to move. If there is still a collision between robots, the incremental delay time $t_0$ is added until there are no conflicts. As for the strategy of assigning fixed priority, the robot with higher priority is required to pass the conflicting position first, and the robot with lower priority passes later with a specified time until all conflicts are solved. According to the two strategies above, the cycle time and the final path of the robot are determined.

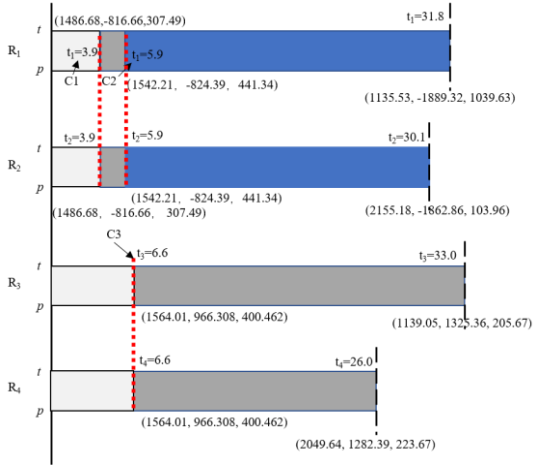

**Fig. 7.** Before multi-robot coordination, Time-position chart based on the principle of proximity (Red dash line represents the moment of conflicts between robots）

**Table 2** Task allocation results based on different methods

| Robot | The principle of proximity | | The proposed local robust allocation method | |
|---|---|---|---|---|
| | *No.* of tasks | Allocated task set | *No.* of tasks | Allocated task set |
| $R_1$ | 12 | $\{M_1, M_2, M_5, M_6, M_7, M_{10}, M_{11}, M_{19}, M_{20}, M_{21}, M_{22}, M_{23}\}$ | 11 | $\{M_1, M_2, M_3, M_7, M_{10}, M_{11}, M_{19}, M_{20}, M_{21}, M_{22}, M_{23}\}$ |
| $R_2$ | 11 | $\{M_3, M_4, M_8, M_9, M_{12}, M_{13}, M_{14}, M_{15}, M_{16}, M_{17}, M_{18}\}$ | 12 | $\{M_4, M_5, M_6, M_8, M_9, M_{12}, M_{13}, M_{14}, M_{15}, M_{16}, M_{17}, M_{18}\}$ |
| $R_3$ | 12 | $\{M_{24}, M_{25}, M_{26}, M_{27}, M_{28}, M_{29}, M_{30}, M_{31}, M_{32}, M_{33}, M_{34}, M_{36}\}$ | 11 | $\{M_{24}, M_{25}, M_{26}, M_{27}, M_{28}, M_{29}, M_{30}, M_{31}, M_{32}, M_{33}, M_{34}\}$ |
| $R_4$ | 10 | $\{M_{35}, M_{37}, M_{38}, M_{39}, M_{40}, M_{41}, M_{42}, M_{43}, M_{44}, M_{45}\}$ | 11 | $\{M_{35}, M_{36}, M_{37}, M_{38}, M_{39}, M_{40}, M_{41}, M_{42}, M_{43}, M_{44}, M_{45}\}$ |

As shown in Fig. 8, conflict D1 was caused by $R_1$ and $R_2$ robots, and D2 was caused by $R_3$ and $R_4$. The conflict D1

was generated during the $R_1$ probe movement from $M_3$ to $M_7$, and at the same time, $R_2$ moved from $M_6$ to $M_8$.

Yinhua Liu[1], Wenzheng Zhao[1], Tim Lutz[2], Xiaowei Yue[2,*]

According to the priority of two robots, the poses of measuring positions of $M_3$ to $M_7$ for the robot $R_1$ are modified to avoid conflicts. The updated probe orientation was searched by satisfying two conditions: (1) it is in the plane passing the nominal vector of $M_3$ and (2) the angle between it and $M_7$ is equal to the angle formed by $M_3$ and $M_7$ in order to find the pose that the robot needs to move after measuring the point $M_3$. Similarly, the D2 conflict between $R_2$ and $R_4$ was eliminated. Fig. 9 is the final motion path diagram based on the proposed local robust task allocation and multi-robot coordination strategy. In Fig. 9, the solid black lines represent the path and probe pose modification results for the conflicts D1 and D2. The proposed method can reduce the conflicts among robots, and at the same time, the coordination method can ensure that the robot path can be modified for conflict elimination automatically without robots' stopping.

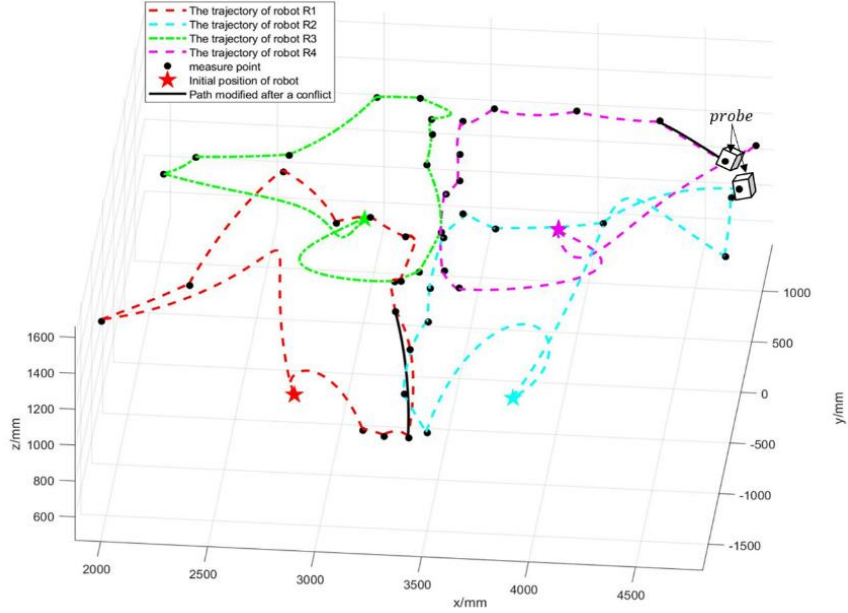

**Fig. 9.** Robot motion path based on the proposed multi-robot coordination strategy (The red dashed line represents the trajectory of the $R_1$; The indigo dashed line represents the trajectory of the $R_2$; The green line represents the trajectory of the $R_3$; The pink dashed line represents the trajectory of the $R_4$; ● represents the measurement point; ★ represents the initial position of the end-effector; — represents the modified path for conflict elimination).

Based on the non-conflict path of robots, the motion time of each path can be estimated. Table 3 showed the comparative results based on different task allocation and coordinated planning strategies. The results of the first two coordination strategies were obtained based on the proximity principle for task allocation. The results of the proposed method were based on the local robust task allocation method and the coordinated strategy. As shown in Table 3, the cycle time of the non-conflict path of multi-robots based on the proposed method was 28.5 seconds, which was reduced by 41.1% and 28.8% compared to the delay startup and fixed priority methods. The maximum time difference $\Delta t$ among the robots is only 1.4 seconds based on the proposed method. The difference was significantly reduced compared to the 11.2s of delay startup method and 7.7s of the fixed priority method.

To evaluate the robustness performance of the proposed method, the MPs were randomly selected from the auto body surfaces. The task allocation, reachability analysis, path planning of each robot and the coordinated strategy were done for multiple cases. The performance of the cycle time and motion time consistency based on different methods were calculated. Since the MP locations on the left and right side of the auto body are nearly symmetrical, only the left side of the auto body with two robots was considered for the 3 cases. Table 4 showed the final results of multi-robot inspection optimization.

Based on Table 4, the cycle time of the multi-robot inspection station and inspection time difference among robots were shown in Fig. 10. On average, the proposed method took less inspection time to complete the tasks than the other two benchmark methods. It means that more MPs can be measured for a given cycle time of an inline assembly line. Besides, the time difference $\Delta t$ of the two robots is less than the other two benchmark strategies. The proposed method decreases some conflicts in the phase of task allocation and reduces the cycle time and motion time difference of all robots to improve the utilization rate of each robot. In addition, because of non-stop of all the robots during conflict elimination, the deadlock phenomena (due to waiting of the lower-priority robots) were reduced significantly. This lays a solid foundation for wide use of

Yinhua Liu[1], Wenzheng Zhao[1], Tim Lutz[2], Xiaowei Yue[2,*]

the autonomous multi-robot inspection system. Although we used the auto body as an example in the case study, the technique can also be extendable and applicable to inspection for aircraft production, such as composite parts assembly (Yue and Shi, 2018) and fuselage assembly (Wen et al., 2019).

**Table 3** Performance comparison based on different task allocation methods

| Robot | Delay in startup | Fixed priority | The Proposed method |
|-------|------------------|----------------|---------------------|
| $R_1$ | 31.8 | 31.8 | 27.5 |
| $R_2$ | 40.2 | 36.7 | 28.5 |
| $R_3$ | 33 | 33 | 28.3 |
| $R_4$ | 29 | 29 | 27.1 |
| Cycle time ($s$) | 40.2 | 36.7 | 28.5 |
| $\Delta t(s)$ | 11.2 | 7.7 | 1.4 |

**Table 4** Results of the cases with differents MP locations

| Case number | Robot | Number of allocated tasks | | Inspection time ($s$) | | |
|-------------|-------|----------------------------|--|------------------------|--|--|
| | | Principle of proximity | Proposed allocation method | Delay setup | Fixed priority | Proposed method |
| 1 | $R_1$ | 13 | 9 | 18.1 | 18.1 | 22.6 |
| | $R_2$ | 7 | 11 | 31.2 | 31.2 | 27.5 |
| 2 | $R_1$ | 12 | 10 | 30.5 | 30.5 | 26.1 |
| | $R_2$ | 9 | 11 | 38.3 | 35.5 | 29.3 |
| 3 | $R_1$ | 11 | 11 | 36.2 | 33.5 | 27.6 |
| | $R_2$ | 11 | 11 | 28.6 | 28.6 | 29.7 |

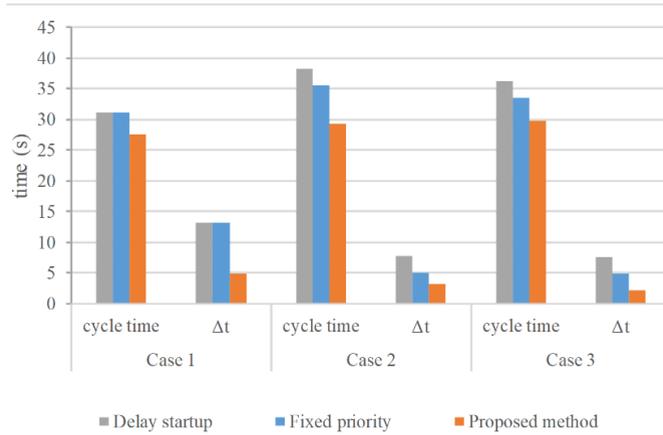

**Fig. 10.** Results comparison of different coordinated motion planning methods

## 6. Summary

Collision-free path planning and multi-robot coordination is an important research area, and it has large impacts on robotics, autonomous systems, smart manufacturing, automotive and aerospace industries, etc. The motion planning in the autonomous multi-robot optical inspection system needs integrated efforts on task assignment, robot path planning, and multi-module coordination. Although numerous studies have been done in this field, existing approaches cannot be applied to our autonomous multi-robot inspection system directly due to two aspects: (1) it is difficult to implement these approaches in complex large-scale free-form surfaces because the optical inspection system has unique requirements on robotic end-effectors' orientations; (2) the optimization computation in existing approaches is time-consuming and cannot satisfy the inline inspection requirement. To realize the multi-robot optical inspection for the inline auto body assembly line, we proposed a new method that integrates local robust task allocation with coordinated motion planning. Firstly, the physical modules of the multi-robot optical inspection systems were introduced in detail, and corresponding constraints and technical assumptions were summarized. To improve inspection efficiency, the tasks with short distances and

Yinhua Liu[1], Wenzheng Zhao[1], Tim Lutz[2], Xiaowei Yue[2,*]

similar probe poses would be allocated to the same robot. A local robust task allocation approach was proposed to reduce inspection time. Collision-free path planning and coordinated motion planning for autonomous multi-robot inspection were proposed to mitigate the risk of collision between robots and environments, resolve conflicts between robots, as well as reduce the cycle time for dimensional quality inspection. The case study shows the proposed method can indeed realize effective task allocation and conflict avoidance. Moreover, the cycle time is significantly reduced compared to benchmark methods.

**Acknowledgments** Dr. Liu's research was partially funded by National Science Foundation of China (51875362) and State Key Laboratory of Mechanical System and Vibration (MSV202010).